\documentclass{ieee-ojvt}
\usepackage{cite}
\usepackage{amsmath,amssymb,amsfonts}
\usepackage{algorithmic}
\usepackage{graphicx,color}
\usepackage{textcomp}
\usepackage{booktabs}
\usepackage{multirow}
\usepackage{balance}
\usepackage{pifont}
\usepackage{tabularx}
\usepackage{ragged2e}
\usepackage{float}
\usepackage{placeins}
\usepackage[T1]{fontenc}
\newcommand{\cmark}{\ding{51}}%
\newcommand{\xmark}{\ding{55}}%
\def\BibTeX{{\rm B\kern-.05em{\sc i\kern-.025em b}\kern-.08em
    T\kern-.1667em\lower.7ex\hbox{E}\kern-.125emX}}
\AtBeginDocument{\definecolor{ojcolor}{cmyk}{0.93,0.59,0.15,0.02}}
\def\OJlogo{\vspace{-12pt}\includegraphics[height=24pt]{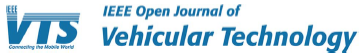}}

\begin{document}
\receiveddate{XX August, 2025}
\reviseddate{XX Month, XXXX}
\accepteddate{XX Month, XXXX}
\publisheddate{XX Month, XXXX}
\currentdate{XX Month, XXXX}
\doiinfo{OJVT.2025.0800000}

\title{Hyperspectral Sensors and Autonomous Driving: Technologies, Limitations, and Opportunities}

\author{
    Imad Ali Shah\textsuperscript{1,2},
    Jiarong Li\textsuperscript{1,2},
    Roshan George\textsuperscript{1,2},
    Tim Brophy\textsuperscript{1,2},
    Enda Ward\textsuperscript{3},
    Martin Glavin\textsuperscript{1,2},
    Edward Jones\textsuperscript{1,2} and 
    Brian Deegan\textsuperscript{1,2}
}

\affil{School of Engineering, University of Galway, Ireland}
\affil{Ryan Institute, University of Galway, Ireland}
\affil{Valeo Vision Systems, Tuam, Ireland}

\authornote{This work has been submitted to the IEEE OJVT for possible publication. Correction and changes are expected, as well as copyright may be transferred to IEEE without notice, after which this version may no longer be accessible.}
\markboth{MANUSCRIPT SUBMISSION (Hyperspectral Sensors and Autonomous Driving: Technologies, Limitations, and Opportunities)}{Shah \textit{et al.}}

\begin{abstract}
Hyperspectral imaging (HSI) offers a transformative sensing modality for Advanced Driver Assistance Systems (ADAS) and autonomous driving (AD) applications, enabling material-level scene understanding through fine spectral resolution beyond the capabilities of traditional RGB imaging. This paper presents the first comprehensive review of HSI for automotive applications, examining the strengths, limitations, and suitability of current HSI technologies in the context of ADAS/AD. In addition to this qualitative review, we analyze 216 commercially available HSI and multispectral imaging cameras, benchmarking them against key automotive criteria: frame rate, spatial resolution, spectral dimensionality, and compliance with AEC-Q100 temperature standards. Our analysis reveals a significant gap between HSI’s demonstrated research potential and its commercial readiness. Only four cameras meet the defined performance thresholds, and none comply with AEC-Q100 requirements. In addition, the paper reviews recent HSI datasets and applications, including semantic segmentation for road surface classification, pedestrian separability, and adverse weather perception. Our review shows that current HSI datasets are limited in terms of scale, spectral consistency, the number of spectral channels, and environmental diversity, posing challenges for the development of perception algorithms and the adequate validation of HSI's true potential in ADAS/AD applications. This review paper establishes the current state of HSI in automotive contexts as of 2025 and outlines key research directions toward practical integration of spectral imaging in ADAS and autonomous systems.
\end{abstract}


\maketitle

\section{Introduction}
\IEEEPARstart{T}{he} rapid advancement of autonomous driving (AD) and Advanced Driver Assistance Systems (ADAS) over the past two decades has fundamentally transformed the automotive industry, promising safer and more efficient transportation systems. Central to these technologies is the ability to accurately perceive and interpret the surrounding environment through various sensor modalities~\cite{grigorescu2020survey}. While traditional perception systems are predominantly based on RGB cameras, LiDAR, and Radar sensors, these conventional approaches face significant limitations in challenging environmental conditions~\cite{vargas2021overview,manivannan2024weather} that compromise vehicle safety and operational reliability.


\begin{figure*}
\centerline{\includegraphics[width=0.99\textwidth]{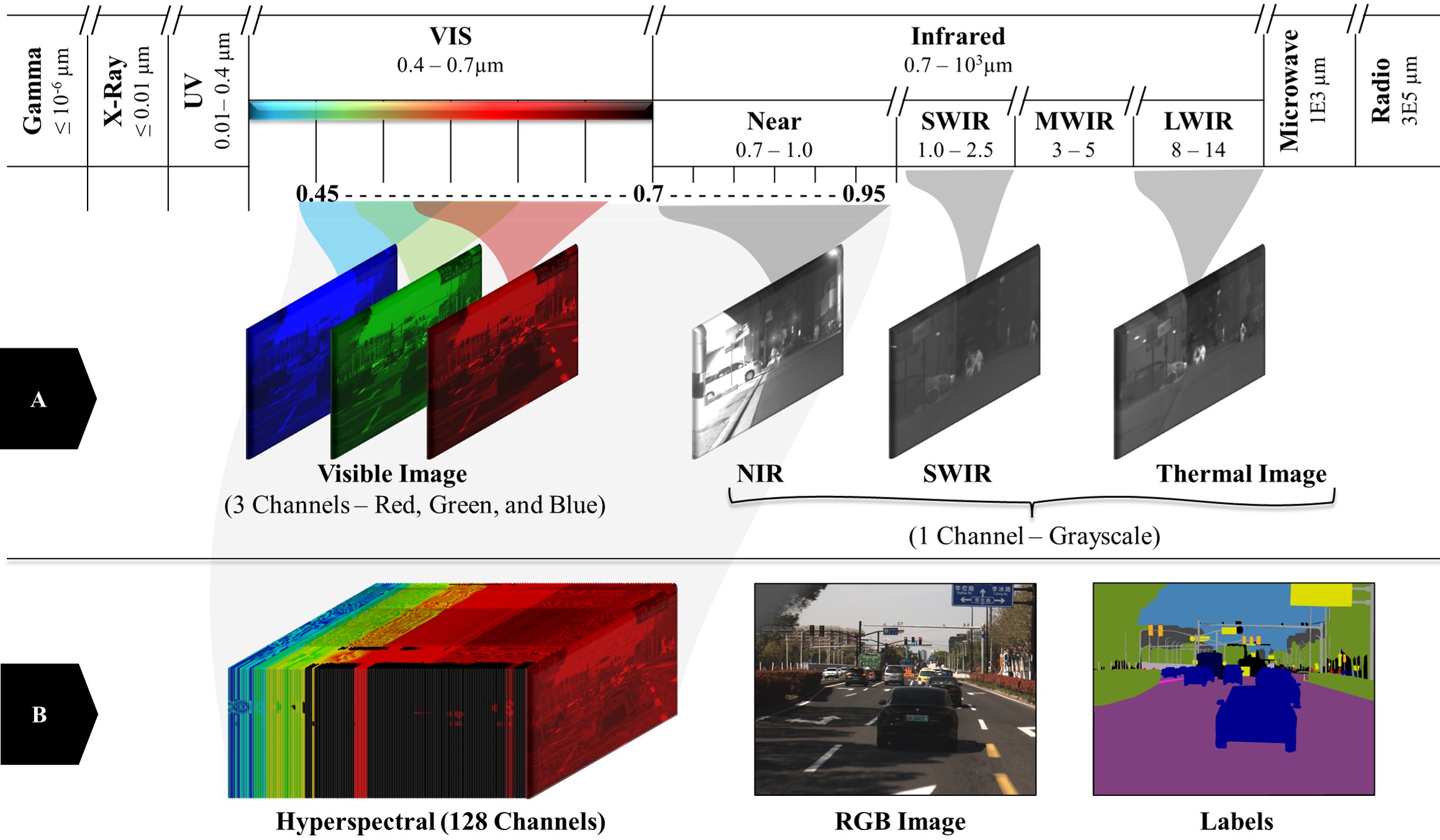}}
\caption{Comparative Overview of Imaging Modalities Across the Electromagnetic Spectrum. (A) Depicts RGB imaging (three channels: red, green, blue) alongside grayscale representations of Near-Infrared (NIR), Short-Wave Infrared (SWIR), and Long-Wave Infrared (LWIR) bands from the Multispectral object detection dataset~\cite{takumi2017multispectral}, spanning the 0.4–14$\mu$m range. Multispectral Imaging (MSI) typically involves combining multiple (around 4-15) broader spectral channels. (B) Illustrates Hyperspectral Imaging (HSI) with 128 narrow, contiguous spectral channels from the Hyperspectral City dataset (0.45–0.95$\mu$m)~\cite{shen4560035urban}, shown with its corresponding RGB image and semantic labels. The figure provides a comparative overview of grayscale, RGB, MSI, and HSI modalities, emphasizing the enhanced spectral resolution of HSI. Note: Color mappings are for representation purposes only.}
\label{label_Fig1.HSIvsRGBvsGS_Overview}
\end{figure*}

Despite the wide adoption of RGB cameras in the current ADAS perception pipeline, they exhibit fundamental constraints when distinguishing materials with similar visual appearances but different compositions~\cite{huang2021weakly}. This phenomenon, known as metamerism, becomes particularly prominent in adverse weather conditions such as fog, rain, snow, and varying lighting scenarios. For example, atmospheric conditions can significantly alter object reflectance in RGB images. Fog causes reflective surfaces to appear washed out due to light scattering~\cite{vargas2021overview}, while wet surfaces from rain or liquid spills can appear darker due to internal light reflections within the liquid medium~\cite{nolet2014measuring}.

Hyperspectral imaging (HSI) is emerging as an enabling technology to address these limitations by capturing electromagnetic (EM) intensities across hundreds of narrow, contiguous spectral bands, as shown in Fig.~\ref{label_Fig1.HSIvsRGBvsGS_Overview}. Unlike conventional RGB cameras that capture only three broad spectral bands, HSI systems can typically operate from visible (VIS: 0.4-0.7$\mu$m) to near-infrared (NIR: 0.7-1$\mu$m) or shortwave infrared (SWIR: 1-2.5$\mu$m) regions, acquiring detailed spectral signatures for precise material identification and environmental characterization~\cite{gutierrez2023chip}. This enhanced spectral resolution enables detection of subtle material differences invisible to RGB cameras and human vision, providing critical advantages for object detection~\cite{lee2021channel}, road surface condition assessment~\cite{valme2024road,ozdemir2020neural,abdellatif2020pavement}, and pedestrian separability~\cite{li2025hyperspectralvsrgbpedestrian} under challenging conditions.

Recent advances in HSI technologies have accelerated HSI research in automotive applications, as these innovations have resulted in compact and video-capable cameras, making real-time deployment increasingly viable for dynamic ADAS/AD environments~\cite{gutierrez2022exploring}. Industry analyses consistently indicate rapid growth in the HSI market, though market valuations vary considerably depending on study scope and methodology. Market assessments for HSI systems and accessories estimate values of approximately USD 0.28-0.85 billion for 2023-2025~\cite{BCCPublishing2025,SkyQuestt2025,MarketsandMarket2025}, while broader market evaluations report significantly higher values of USD 14-16 billion for 2023-2024~\cite{GrandViewResearch2025,StraitsResearch2025}. Across all of these reports, the projected compound annual growth rates are 8-15\%, reflecting both substantial commercial interest and rapid technological advancement in HSI. However, these promising commercial prospects have not translated to widespread adoption of HSI in ADAS/AD systems. Persistent challenges, including high snapshot camera costs, demanding data throughput requirements, and complex system integration, have collectively hindered both research focus and practical implementation of HSI in automotive environments.


\begin{figure*}[h!]
\small
\begin{tabular}{l}  
     \hspace{1.3cm} Section-\ref{Section:HSI_Fundamentals} \hspace{2.6cm} Section-\ref{Section:Current_HSI_State} \hspace{3cm} Section-\ref{Section:HSI_Availability_Suitability} \hspace{3.2cm} Section-\ref{Section:Future_Directions}
\end{tabular}
\\
\centerline{\includegraphics[width=0.999\textwidth]{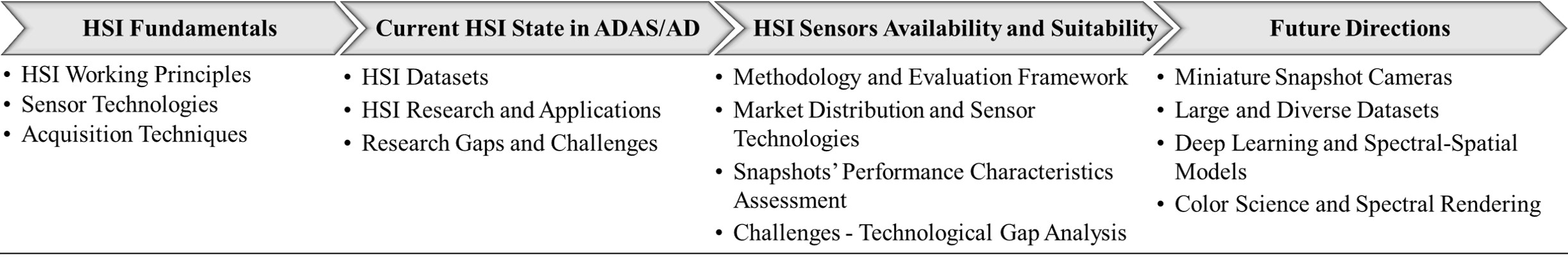}}
\caption{Review paper structure: Outlining HSI fundamentals, current state, sensor considerations, and future research opportunities for ADAS/AD.}
\label{label_Fig0.PaperOrganization}
\end{figure*}

While HSI research remains limited in automotive applications compared to other domains such as remote sensing and medicine, where HSI is widely adopted, HSI has started to demonstrate its application in driving scenarios, including perception capability~\cite{shah2024hyperspectral}, road surface condition classification~\cite{valme2024road}, and terrain recognition (e.g., grass, bumpy road, asphalt)~\cite{jakubczyk2022hyperspectral}. The development of specialized HSI datasets has begun establishing foundations for automotive applications, with some recent contributions including HyKo~\cite{winkens2017hyko}, HSI-Drive~\cite{gutierrez2023hsi}, Hyperspectral City~\cite{shen4560035urban} (H-City), and HSI-Road~\cite{lu2020hsi} datasets. These datasets are becoming foundational for enabling algorithm development and validation in ADAS/AD. However, there are significant gaps remaining in their data diversity, spectral and spatial resolution, annotation quality, and driving scene coverage~\cite{shah2024hyperspectral}.

Despite promising developments, HSI adoption in ADAS/AD faces substantial challenges. The high dimensionality of hyperspectral data can be many times larger than RGB (e.g., 3-channel RGB vs 128 channels of the H-City dataset), posing significant computational burdens for real-time processing~\cite{bhargava2024hyperspectral}. Current snapshot cameras trade spectral resolution for temporal capability and spatial resolution. Integration challenges include sensor synchronization with existing automotive architectures, data storage requirements, and the need for specialized processing hardware. However, the main bottleneck is the lack of automotive-grade hyperspectral sensors and industry standardization, which impedes both research activities and commercial deployment. Addressing these challenges requires a comprehensive understanding of the current HSI technology landscape in ADAS/AD.

To provide an overview of the current state of HSI technology as a first reference guide for automotive researchers and industry specialists, this paper presents a comprehensive review of HSI technology for ADAS/AD applications. This review is the first to systematically evaluate over 216 commercially available HSI and MSI cameras, addressing a critical gap in the literature and assessing their suitability for automotive deployment.

We begin by establishing the fundamentals of HSI technology (Section~\ref{Section:HSI_Fundamentals}) relevant to automotive applications, including key principles, spectral imaging concepts, and the fundamental trade-offs between snapshot- and scanning-based HSI architectures and their implications for real-time automotive performance.

We then investigate the current landscape of HSI in ADAS/AD research (Section~\ref{Section:Current_HSI_State}), examining existing datasets and identifying critical gaps that limit algorithm development.

Subsequently, we conduct a comprehensive evaluation of commercially available HSI sensors (Section~\ref{Section:HSI_Availability_Suitability}), assessing their specifications based on critical parameters determining vehicle integration viability, including acquisition types, operating temperature, spectral coverage, spatial and spectral resolutions, frame rates, weight, and power consumption.

Through examination of integration challenges, sensor limitations, and technical constraints, we provide insights into factors constraining HSI adoption in the automotive sector. By establishing a comprehensive baseline of HSI technology capabilities as of 2025, this review offers guidance for researchers and industry practitioners advancing spectral imaging applications in ADAS/AD.

The remainder of this paper is organized as shown in Fig.~\ref{label_Fig0.PaperOrganization} and concludes in Section~\ref{Section:Conclusion} with key findings and recommendations to advance HSI technology in ADAS/AD context.

\section{HSI Fundamentals and Technical Overview}
\label{Section:HSI_Fundamentals}
This section provides an overview of HSI fundamentals, covering the core principles, sensor technologies, and acquisition techniques that enable HSI systems to capture detailed spectral information. Understanding these foundational concepts is essential for evaluating HSI's potential applications in ADAS/AD scenarios.

\begin{figure*}[h!]
\centerline{\includegraphics[width=0.99\textwidth]{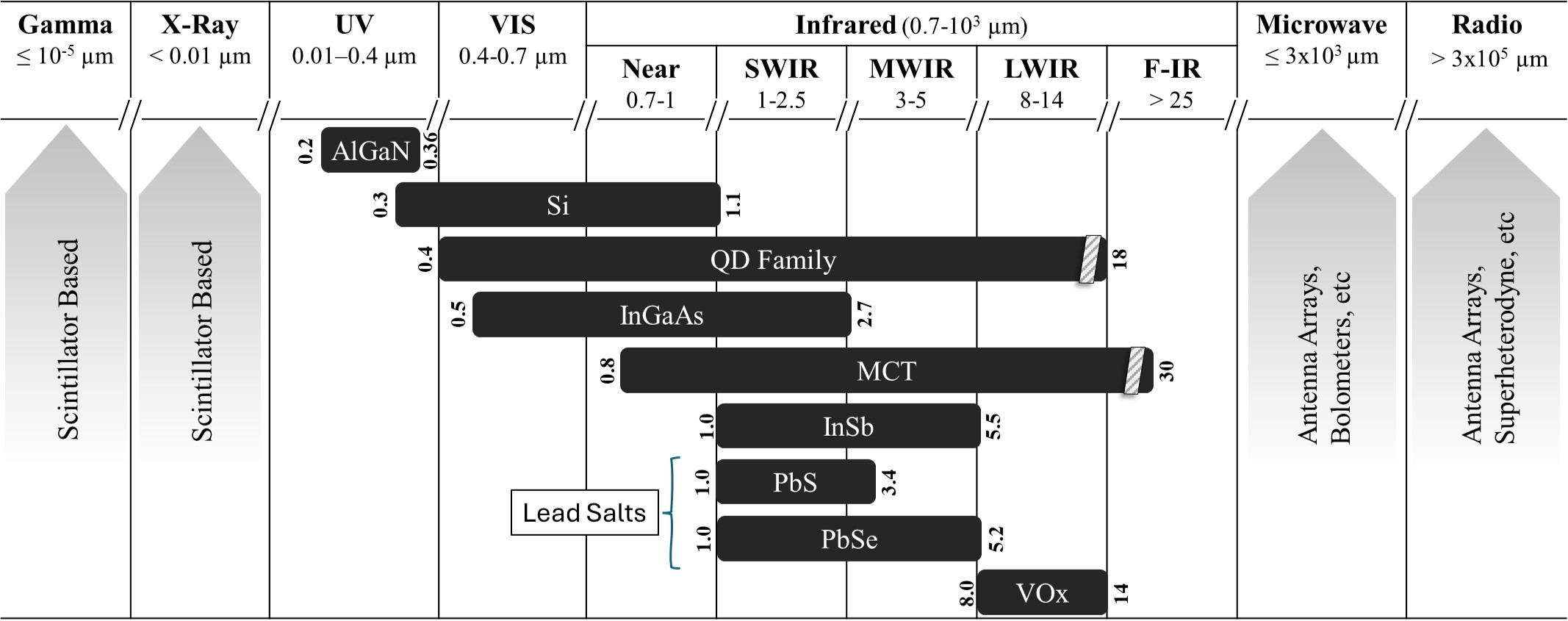}}
\caption{Overview of sensors (or detectors), and their EM ranges. According to our review of over 216 cameras, discussed in Section~\ref{Section:HSI_Availability_Suitability}, more than 87$\%$ of commercially available HSI cameras utilize Si and InGaAs-based sensors. Note that the boundaries of these sensors can have variations of $\pm$0.1- 0.3$\mu$m, as reported by different researchers and industrial manufacturers. Details of each sensor are discussed in Section~\ref{Section:HSI_Fundamentals}(\ref{Section:HSI_Fundamentals-SUB:Sensor Technologies}).}
\label{label_Fig.SensorsDetectors}
\end{figure*}

\subsection{HSI Working Principles}
\label{Section:HSI_Fundamentals-SUB:HSI Working Principles}
Hyperspectral-based digital imaging leverages spectroscopy principles by studying how EM radiation interacts with matter. This interaction involves phenomena such as absorption, transmission, reflection, and emission~\cite{falcioni2023assessment}, which vary across the EM spectrum in ways unique to each material's composition. HSI captures this spectral information across a broad range of the EM spectrum, utilizing hundreds of narrow and contiguous spectral bands~\cite{goetz1985imaging} that span from the ultraviolet (UV) through the visible (VIS) and into the infrared (NIR, SWIR, etc.) regions (typically 0.4-2.5$\mu$m~\cite{bhargava2024hyperspectral}).

The output of an HSI system is a three-dimensional data structure, commonly known as a hypercube, as depicted in Fig.~\ref{label_Fig1.HSIvsRGBvsGS_Overview}, which comprises two spatial dimensions (representing width and height) and a third spectral dimension of the imaged scene. Each pixel within the spatial dimensions contains a contiguous spectrum that measures light intensity at each captured wavelength, forming the spectral signature of the corresponding spatial point in the captured scene.

This rich per pixel spectral information fundamentally distinguishes HSI from other imaging modalities. Traditional RGB imaging captures information in only three broad and overlapping spectral bands (red, green, and blue), while multispectral imaging (MSI) captures data in a few discrete spectral bands, often with gaps between them, and typically ranging from 3 to 15~\cite{mengu2023snapshot}, though definitions vary among researchers. The key advantage of HSI lies in its high spectral resolution, characterized by narrow and contiguous spectral bands that enable fine spectral sampling for accurate material differentiation.

This fine spectral resolution enables HSI to overcome limitations inherent in conventional imaging approaches. For instance, it can mitigate the metamerism effect, where visually similar objects appear identical in standard RGB imaging~\cite{shen4560035urban}. Different materials possess distinct spectral signatures that act like chemical fingerprints, determined by their unique chemical and physical properties. This capability allows HSI to distinguish between different vegetation types~\cite{ye2024mangrove}, assess plant health~\cite{qamar2022impacts} based on spectral reflectance patterns, or identify ice on road surfaces through characteristic spectral signatures~\cite{valme2024road}. While both HSI and MSI can capture data beyond the visible spectrum, HSI's contiguous and narrow bands provide significantly richer data, detecting subtle spectral features that MSI sensors with fewer and broader bands might miss, thereby enabling more precise material identification and classification.

\begin{table}[h!]
\centering
\caption{Suitability of Imaging Sensors: EM Spectrum (in $\mu$m)}
\label{tab:sensor_comparison}
\resizebox{\columnwidth}{!}{%
\begin{tabular}{@{}lcccccccc@{}}
\toprule
\textbf{Sensor} & \textbf{0.4$\sim$0.8} & \textbf{$<$1.1} & \textbf{$<$1.7} & \textbf{$<$3} & \textbf{$<$5} & \textbf{$<$8} & \textbf{$<$14} & \textbf{14+}\\
\midrule
\textbf{Si} & \textcolor{green}{\cmark} & \cmark & \cmark & \textcolor{red}{\xmark} & \textcolor{red}{\xmark} & \textcolor{red}{\xmark} & \textcolor{red}{\xmark}  & \textcolor{red}{\xmark}\\
\textbf{QD} & \cmark & \cmark & \cmark & \cmark & \cmark & \cmark & \cmark & \cmark\\
\textbf{InGaAs} & \cmark & \textcolor{green}{\cmark} & \textcolor{green}{\cmark} & \cmark & \textcolor{red}{\xmark} & \textcolor{red}{\xmark}  & \textcolor{red}{\xmark}  & \textcolor{red}{\xmark}\\
\textbf{MCT} & \textcolor{red}{\xmark} & \cmark & \cmark & \cmark & \cmark & \textcolor{green}{\cmark}  & \cmark & \textcolor{green}{\cmark} \\
\textbf{InSb} & \textcolor{red}{\xmark} & \textcolor{red}{\xmark} & \textcolor{red}{\xmark} & \cmark & \cmark\textsuperscript{*} & \textcolor{red}{\xmark}  & \textcolor{red}{\xmark}  & \textcolor{red}{\xmark}\\
\textbf{Lead Salts} & \textcolor{red}{\xmark} & \textcolor{red}{\xmark} & \textcolor{red}{\xmark} & \textcolor{green}{\cmark} & \textcolor{green}{\cmark} & \textcolor{red}{\xmark}  & \textcolor{red}{\xmark}  & \textcolor{red}{\xmark}\\
\textbf{VOx} & \textcolor{red}{\xmark} & \textcolor{red}{\xmark} & \textcolor{red}{\xmark} & \textcolor{red}{\xmark}  & \cmark & \cmark & \textcolor{green}{\cmark}  & \textcolor{red}{\xmark}\\
\bottomrule
\end{tabular}%
}
\parbox{\columnwidth}{
\footnotesize
\quad \\
\textcolor{green}{\cmark}: Suitable, \quad \cmark: Applicable, \quad \textcolor{red}{\xmark}: Not Applicable \\
* Better sensitivity and faster response than Lead Salts, but requires cooling
}
\end{table}

\subsection{Sensor Technologies}
\label{Section:HSI_Fundamentals-SUB:Sensor Technologies}
The performance of any camera relies on its sensor, which ideally captures data with high sensitivity, low noise, and fast readout in spectral, spatial, and temporal dimensions~\cite{bhargava2024hyperspectral}. The robustness of HSI, being more complex (i.e., capturing hundreds of subtle spectral variations), is relatively more dependent on the type of image sensor than traditional RGB. Our reviewed sensors for EM spectrum bands are shown in Table~\ref{tab:sensor_comparison} and Fig.~\ref{label_Fig.SensorsDetectors}, providing an overview of their suitability, limitations, and spectral ranges. These sensors cover from UV up to long-wave infrared (LWIR) regions.

\textbf{Silicon (Si)} remains the backbone for imaging in VIS and NIR~\cite{huang2006microstructured,chen2022recent} (VNIR, 0.3–1.1$\mu$m) regions, due to their mature technology and high pixel counts. Their cost-effectiveness and well-established manufacturing processes make them suitable for many ADAS/AD applications, particularly in daylight conditions. However, their response falls off rapidly after $\sim$0.8$\mu$m, and their indirect bandgap fundamentally limits its usability beyond $\sim$1.0$\mu$m. This limits its viability in adverse weather conditions, which are required in urban driving-based dynamic scenarios. Therefore, the limitation of Si to VNIR requires dependence on other materials for applications requiring sensing of other EM spectra.

\textbf{Indium Gallium Arsenide (InGaAs)} sensors typically cover SWIR~\cite{martin2005320x240,chen2022recent} region and offer sensitivity from 0.9 to 1.7$\mu$m in standard~\cite{gong2009near}, up to 2.6$\mu$m in extended‐cutoff~\cite{arslan2015extended}, and 0.4$\sim$1.7$\mu$m in VIS-extended~\cite{bhargava2024hyperspectral} based InGaAs variants. The spectral range of InGaAs can be extended using tunable components, such as Indium phosphide (InP) substrate, providing 0.9-2.65$\mu$m spectral response~\cite{arslan2015extended}, and dual-channel detection (e.g., VNIR) by the use of multi-layered Molybdenum disulfide (MoS$_2$) and Indium Aluminum Arsenide (InAlAs)~\cite{deng2019integration,kang2019inas}. The family of InGaAs variants covering from NIR to SWIR regions is quite large~\cite{sun2010design}, with a wide InGaAs-based variants proposed for different ranges. These extended ranges, however, come with a higher cost and active cooling requirements. Nonetheless, these limitations are actively being researched, and recent uncooled-based InGaAs with mosaic filters have been introduced~\cite{blanch2022compact}, enabling SWIR-based spectral imaging opportunities. 

\textbf{Quantum Dots (QD)} is an emerging technology for broad detection, with colloidal QD (CQD) promising room temperature operation across a wide tunable bands from $\sim$0.4 up to 18$\mu$m~\cite{wang2024extending}. Using nanocrystals based on lead-selenide (PbSe), lead-sulfide (PbS), or mercury-telluride (HgTe), CQDs can achieve continuous coverage from 0.5 to 4.0$\mu$m~\cite{sun2024innovative}. Similar to InGaAs-variants, the QD's family is also quite large, each with their own spectral coverage, external quantum efficiency (EQE), detectivity, etc., and due to multiple exciton generation, the EQE can reach up to 100\%~\cite{guo2022advances}. Though still under active development, QDs promise monolithic, low‐cost, large‐area focal plane arrays, and their tunability with high resolution across multiple spectra makes them a promising candidate for advanced sensor systems and ADAS/AD applications. Furthermore, their high sensitivity and low dark current~\cite{chen2025hybrid} could facilitate object detection and classification in complex urban environments.

\textbf{Indium Antimonide (InSb)} and \textbf{Mercury Cadmium Telluride (HgCdTe, or MCT)} predominantly work in the midwave infrared (MWIR, $\sim$3–5$\mu$m)~\cite{zhang2011long} to LWIR ($\sim$8–14$\mu$m)~\cite{chen2021mechanism} regions~\cite{chen2022recent}. Where InSb detectors (1.0–5.5$\mu$m) offer uniformity and low noise in MWIR under moderate cooling~\cite{gunapala2021high,shkedy2021hot}, and MCT covers an extraordinary continuous span up to 25$\mu$m~\cite{wang2024extending} through compositional tuning of the Mercury/Cadmium ratio~\cite{rogalski2005hgcdte}, and can be extended up to 30$\mu$m. However, these materials suffer from lattice mismatch-induced defects and require active (such as cryogenic and thermoelectric) cooling~\cite{sun2024innovative}.

The uncooled \textbf{Vanadium Oxide (VO$_\text{X}$)} microbolometers primarily operate in the LWIR (8–14$\mu$m) region, as used in FLIR cameras~\cite{honniball2017spectral}, through temperature-dependent resistance changes in VO$_\text{X}$ films. They are lightweight with low power consumption and their capability to operate at room temperature~\cite{yu2020low}, making them suitable for portable applications~\cite {wang2012vanadium} similar to ADAS/AD, such as VO$_\text{X}$ night-vision and pedestrian detection in challenging conditions~\cite{petilli2023monocular}.

Lead salts, i.e, \textbf{Lead Selenide (PbSe)} and \textbf{Lead Sulfide (PbS)}, were the first detectors used for IR (SWIR to MWIR: 1-5$\mu$) sensing~\cite{gupta2021photoconductive}. PbSe sensor covers 1–4.7$\mu$m and the cooled sensor can extend up to 5.2$\mu$m, whereas the uncooled PbS covers 1-3$\mu$m and up to 3.4$\mu$m when cooled~\cite{PbSeriesSensors}. The viability of a sensor in the MWIR region requires operating in the cryogenic temperature range to prevent thermal generation of charge carriers, making these lead salts a preferred choice over other sensors such as InSb and MCT~\cite{gupta2021photoconductive}. Thus, they are widely used in CQD for more affordable options to cover the SWIR to MWIR regions.
 
Although a broad range of sensors exists, such as the use of AlGaN~\cite{10.1063/1.123358,photonics_algan_camera} for UV imaging (0.2$\sim$0.36$\mu$m), Si and InGaAs families remain the dominant sensors in commercial HSI cameras due to their performance, reliability, and supply-chain maturity.

\begin{figure}
\centerline{\includegraphics[width=0.49\textwidth]{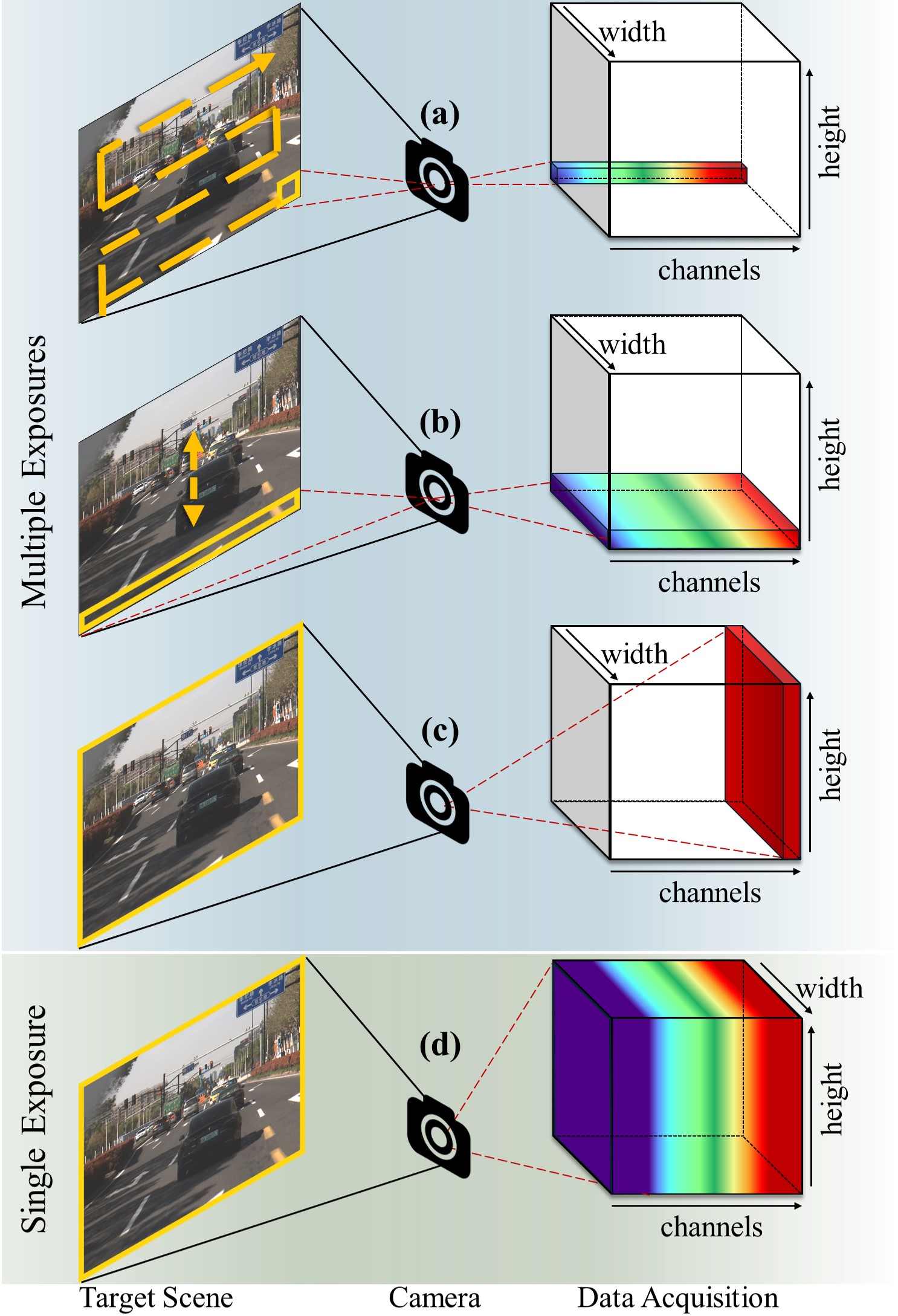}}
\caption{Overview of HSI acquisition methods where yellow lines show the captured spatial location and Violets-to-Red gradient represents the spectral information captured by the respective method: (a) Point or Whiskbroom, (b) Line or Pushbroom, (c) Wavelength Scan using filter wheel or tunable filters, and (d) Snapshot or Global Shutter.}
\label{label_Fig2.HSIAquisitionTypes}
\end{figure}

\subsection{Acquisition Techniques}
\label{Section:HSI_Fundamentals-SUB:Acquisition Techniques}
HSI systems capture a three-dimensional hypercube, comprising two spatial and one spectral dimensions, using various acquisition techniques: spatial scanning, spectral scanning, and snapshot, illustrated in Fig~\ref{label_Fig2.HSIAquisitionTypes}, as well as some hybrid methods. These methods differ in how they sample spatial and spectral information, each with trade-offs in its acquisition speed, spectral and spatial resolutions, which directly affect their suitability for ADAS/AD applications, where real-time performance and dynamic scene handling are critical:

\subsubsection{Spatial Scanning Camera}
Spatial scanning cameras involve moving the sensor (or target) to capture the hypercube slice by slice, widely used in remote sensing~\cite{fowler2014compressive}, but is less ideal for dynamic ADAS/AD scenarios due to potential motion artifacts:
\begin{itemize}
    \item \textbf{Whiskbroom (or PointScan)}: Captures the full spectrum of a single spatial point at a time and has long acquisition times for full spatial scene capture. This makes it unsuitable for ADAS/AD applications where dynamic scene objects (i.e., vehicles, pedestrians, and traffic control signs, etc.) must be captured. These scanners also have fast-moving parts, which makes them prone to high degradation, due to which Linescan cameras were adopted by the succeeding family of Landsat 7 satellite~\cite{NASA_Report2017}.
    \item \textbf{Pushbroom (or Linescan)}: Captures the full spectrum of an entire spatial line of pixels at a time, then moves perpendicularly to capture the full scene. While relatively faster than whiskbroom, it still requires extended acquisition times that can cause object shape distortions in dynamic driving scenes. For instance, Specim IQ can take up 0.5$\sim$256 seconds~\cite{SpecimIQ_IT_Video}, depending on the integration time of 1$\sim$500 milliseconds (ms)~\cite{SpecimIQ_Konica_Ref} to construct a 512x512 spatial image. Although some research reports achieving over 40,000 lines/s~\cite{Vision_Systems2024}, it still requires mechanical scanning for complete spatial coverage. For a 1024 spatial lines-based image, this system will produce a maximum of $\sim$39 frames per second at its full capacity.
\end{itemize}

\subsubsection{Spectral (or Wavelength) Scanning  Camera}
Spectral scanning cameras capture the entire spatial scene at once but acquire one spectral band at a time. Sequential capture causes fast-moving objects to be recorded at different spatial locations when switching between spectral filters, resulting in edge blurring, and can also corrupt the spectra of closely adjacent objects:
\begin{itemize}
    \item \textbf{Tunable filters (TF)}: Methods like Liquid crystal TF~\cite{LCTF_Report} use electronically controlled liquid crystal elements to transmit selectable wavelengths, whereas Acousto-optic TF are solid-state devices~\cite{xu1992modulators} with no moving parts~\cite{korablev2015development,dekemper2012tunable} that provide dynamic spectral band control for fast and precise transmissive TF~\cite{mi2024thermal}. Such systems capture spectral bands sequentially, making them slower and prone to motion artifacts~\cite{batshev2021spectral}.
    \item \textbf{Linear Variable Filters (LVF)} are interference filters with spatially varying transmission wavelengths along their length~\cite{wang2025infrared}, allowing different spectral bands to be captured by positioning detectors or scanning the filter. While more compact than traditional TF, they still have the inherent sequential acquisition, which can lead to motion artifacts in dynamic scenes, and require a spatial-spectral trade-off, limiting their imaging capabilities. 
\end{itemize}

\subsubsection{Snapshot Camera}
Snapshot cameras capture the entire hypercube in a single exposure, ideal for ADAS/AD-based dynamic scenes by enabling real-time capture without scanning. However, they are constrained by the trade-offs between spectral, spatial, and temporal resolutions~\cite{fotiadou2016spectral,gutierrez2023hsi}, a challenge for capturing fast-moving objects under varying lighting conditions with high spectral and spatial resolutions:
\begin{itemize}
    \item \textbf{Mosaic Filter Arrays}: Captures the scene using a grid of multiple spectral filters on pixels, similar to Bayer filters in color cameras. They offer high frame rates suitable for dynamic scenes, but with spatial-spectral resolution trade-offs. As the number of filters increases, spatial resolution reduces. For example, Geelen et al.~\cite{geelen2014compact} demonstrated a compact snapshot imager providing high spatial resolution with 1-4 spectral filters or lower spatial resolution with up to 36 filters in 6×6 arrangements.
    \item \textbf{Integral Field Spectroscopy}: Uses lenslet arrays, or fiber bundles, to divide the image into segments for spectral dispersion. Fabry-Perot interferometer-based imagers can achieve up to 100 spectral channels~\cite{luthman2017fluorescence}, but require complex optical components, limiting automotive applications due to robustness constraints.
    \item \textbf{Computational Imaging}: Methods like Coded Aperture Snapshot Spectral Imager~\cite{arguello2013fast} use coded apertures to multiplex spectral data onto 2D sensor, then computationally reconstruct the hypercube. This approach offers high spectral resolution but requires significant processing power and introduces reconstruction latency.

    \item \textbf{Beamsplitting Systems}: Use multiple sensors with different spectral filters to simultaneously capture various EM spectrum regions~\cite{stoffels1978color,spiering1999multi}, for example, a combination of VIS and SWIR regions. The increase in the number of prisms offers parallel spectral acquisition, but increases system complexity, cost, and noise due to additional optical components~\cite{thomas2025trends}.

    \item \textbf{Other techniques}: Other methods, which are not as commercialized as their counterparts, due to their optical complexity or comparatively lower performance~\cite{thomas2025trends}, include spectrometry-based imaging (i.e., Computed Tomography, Image-Replicating, Image-Mapping, Fourier-Transform, etc)~\cite{descour1996throughput,pawlowski2019high, Cao2021,thomas2025trends}, Multispectral Sagnac interferometer~\cite{kudenov2010white}, and vertically stacked photodiodes~\cite{chen2023bioinspired}.
\end{itemize}

\begin{figure*}[h!]
\centerline{\includegraphics[width=0.99\textwidth]{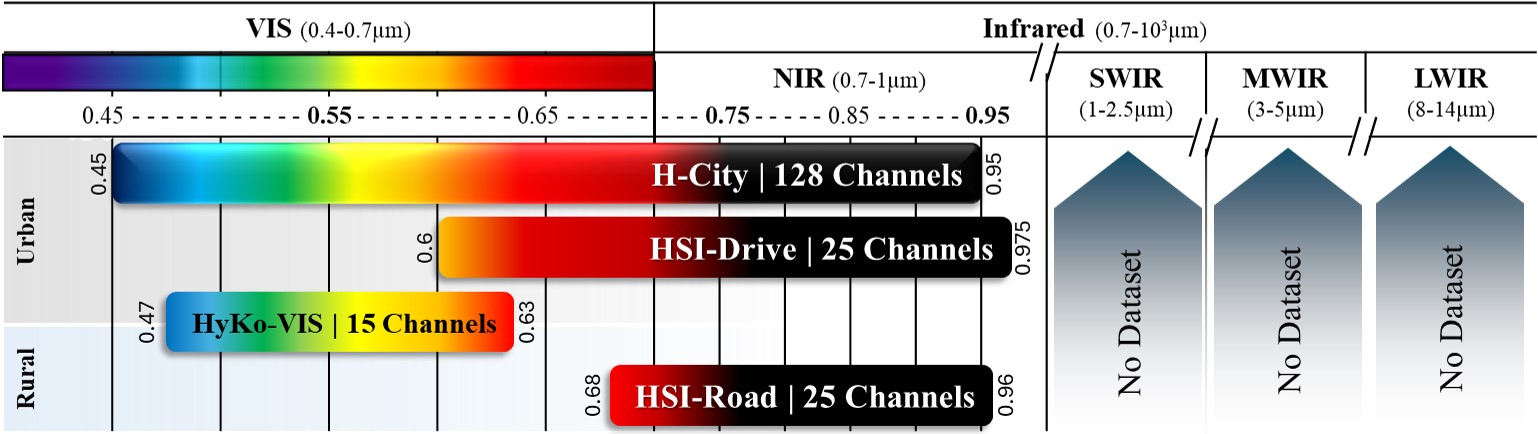}}
\caption{Spectral coverage and channel distribution of available snapshot-based HSI datasets for urban and rural driving scenarios. The visualization highlights each dataset’s coverage in the VIS and NIR regions. Currently, no snapshot-based HSI datasets extend beyond VIS and NIR.}
\label{label_Fig3.HSI_Datasets}
\end{figure*}

\subsubsection{Hybrid Acquisition}
These systems combine multiple imaging modalities to overcome individual camera limitations~\cite{xiong2017snapshot}, such as pairing high-speed low-resolution cameras with low-speed high-resolution cameras~\cite{ben2003motion}, or combining low-resolution HSI with high-resolution RGB~\cite{ma2014acquisition}. Advanced methods like Imec’s Snapscan~\cite{SnapScan_Imec_Int,SnapScan_Imec_IMVE} use internal scanning with striped spectral filters, achieving acquisition times around 200 ms. However, this remains insufficient for high-speed ADAS/AD applications, for instance 30$\sim$60 frame rate-based application~\cite{Lai2022} requires within 16$\sim$33ms of acquisition time.

For ADAS/AD scenarios, snapshot HSI is preferred to avoid motion artifacts, as scanning-based methods are not suited for dynamic scenes containing fast-moving objects and varying illumination~\cite{xiong2017snapshot}. RGB cameras used for ADAS/AD typically operate at 30$\sim$60 frames per second~\cite{Lai2022}, depending on individual application, establishing the baseline performance requirement for HSI systems. Current snapshot systems still face challenges, including spatial-spectral-temporal resolution trade-offs, high costs, and size constraints. Continued advancement toward sub-millisecond integration times~\cite{The_Snapshot_Advantage}, reduced costs, and improved size, weight, and power characteristics are essential for widespread automotive integration.

The choice of sensor technology and acquisition method critically influences the quality and type of hyperspectral data obtained, which directly impacts the design and performance of perception algorithms in ADAS/AD systems. Given the real-time constraints and dynamic complexities of driving scenes, snapshot HSI technologies with optimal spatial, spectral, and temporal resolution trade-offs are essential. Building on these technical foundations, Section~\ref{Section:Current_HSI_State} reviews the current state of HSI adoption

\begin{table*}[h!]
\centering
\caption{Current State of HSI in ADAS/AD Scenarios: Summary of Methodologies used on Publicly Available and Urban HSI Datasets.}
\begin{tabularx}{\textwidth}{@{} p{3.15cm} p{0.5cm} p{4.3cm} >{\noindent\justifying\arraybackslash}X @{}}
\toprule
\textbf{Author} & \textbf{Year} & \textbf{Dataset Used} & \textbf{Method and Techniques} \\
\midrule
Basterretxea et al.~\cite{basterretxea2021hsi} & 2021 & HSI-Drive v1 & Per-pixel ANN followed by two-staged spatial regularization \\
Gutiérrez-Zaballa et al.~\cite{gutierrez2022exploring} & 2022 & HSI-Drive v1 & Patches based UNet, and Per-pixel ANN \\
Gutiérrez-Zaballa et al.~\cite{gutierrez2023hsi} & 2023 & HSI-Drive & Patches based UNet~\cite{ronneberger2015u} \\
Ding et al.~\cite{ding2023dual} & 2023 & H-City & HSI and pseudo-RGB based spectral feature fusion and multi-scale decoder \\
Arnold et al.~\cite{arnold2024spectralzoom} & 2024 & HyKo-VIS & Wandering Patches with Attention Mechanism and ViT-based Segmentation \\
Theisen et al.~\cite{theisen2024hs3} & 2024 & HyKo-VIS, HSI-Drive, and H-City & RGB-HSI benchmarking framework using DeepLabv3+~\cite{chen2017deeplab} and UNet \\
Shah et al.~\cite{shah2024hyperspectral} & 2024 & Relabeled HyKo-VIS, HSI-Drive, and H-City to 6--8 classes, and HSI-Road & DeepLabv3+, HRNet~\cite{sun2019high}, PSPNet~\cite{zhao2017pyramid}, and UNet, along with two Attention Mechanism-based UNet Variants, i.e., CBAM~\cite{woo2018cbam} and CA~\cite{dang2021coordinate} \\
Shah et al.~\cite{shah2025multiscalespectralattentionmodulebased} & 2025 & HyKo-VIS, HSI-Drive, and H-City & Multi-scale spectral feature extraction module, integrated into the U-Net\\
Li et al.~\cite{li2025csnrjmimbasedspectral, li2025hyperspectralvsrgbpedestrian} & 2025 & H-City & CSNR and JMIM-based dimensionality reduction to generate pseudo-RGB, and evaluation through UNet, Deeplabv3+, and Segformer~\cite{xie2021segformer}\\
\bottomrule
\end{tabularx}
\label{tab:LitAndWorkOn_HSI}
\end{table*}

\section{Current State and Use Cases in ADAS/AD}
\label{Section:Current_HSI_State}
With the fundamental principles and enabling hardware of HSI outlined in Section~\ref{Section:HSI_Fundamentals}, this section reviews its current application landscape within ADAS/AD. We examine existing datasets, recent advances in semantic segmentation and material discrimination, and highlight critical gaps and challenges in adopting HSI for automotive perception.

While RGB-based semantic segmentation has established benchmarks through datasets (such as Cityscapes~\cite{cordts2015cityscapes}, KAIST~\cite{hwang2015multispectral}, KITTI~\cite{geiger2012we}, and nuScenes~\cite{caesar2020nuscenes}), HSI-based perception for ADAS/AD remains an emerging field. Recent advances in portable HSI technology, particularly snapshot cameras, have enabled real-time acquisition in dynamic driving scenarios~\cite{gutierrez2022exploring}. The rich spectral information embedded in HSI data presents significant potential for enhanced object classification and material identification beyond the capabilities of conventional RGB-based systems.

\subsection{HSI Datasets}
\label{sec:Current State - HSI Datasets}
\noindent Several specialized HSI datasets have been developed for ADAS/AD applications, each with distinct spectral and spatial characteristics as illustrated in Fig.~\ref{label_Fig3.HSI_Datasets}. Current publicly available HSI datasets are limited to visible (VIS) to near-infrared (NIR) spectral ranges

\begin{itemize}
    \item \textbf{HyKo}~\cite{winkens2017hyko}: Comprises two sub-datasets with HyKo-NIR covering 0.63-0.975~$\mu$m and HyKo-VIS spanning 0.47-0.63~$\mu$m. HyKo-VIS provides 163 hypercubes with 15 spectral bands, while HyKo-NIR contains 78 hypercubes with 25 bands. However, the NIR camera was configured with a narrow, road-focused view that captures the road surface directly rather than forward-facing scenes, which limits its utility for comprehensive scene understanding in ADAS applications.
    \item \textbf{HSI-Drive}~\cite{basterretxea2021hsi,gutierrez2023hsi}: Captures driving scenarios across four seasons and diverse environmental conditions with 25 spectral bands (0.6-0.975~$\mu$m). However, the dataset suffers from coarse pixel-level annotations and limited spectral coverage focused primarily on red and NIR regions, as illustrated in Fig.~\ref{label_Fig3.HSI_Datasets}.
    \item \textbf{H-City}~\cite{you2019hyperspectral,shen4560035urban}: Offers the highest spectral resolution with 128 channels and detailed annotations for 19 object classes, providing fine-grained semantic labels suitable for urban scene understanding. H-City provides the highest spectral resolution (128 channels) and spatial resolution (1422×1889 pixels) among available datasets.
    \item \textbf{HSI-Road}~\cite{lu2020hsi}: Provides binary road surface classification with 25 spectral channels covering the red-NIR region (0.68-0.96~$\mu$m), specifically useful for traversability assessment.
\end{itemize}

Despite these contributions, fundamental challenges persist across existing HSI datasets for ADAS/AD applications~\cite{shah2024hyperspectral}:
\begin{itemize}
\item Lack of standardization in spatial resolution and number of spectral channels
\item Inconsistent spectral coverage limiting cross-dataset generalization
\item Varying annotation schemes and class definitions
\item Significant class-imbalance affecting reliable perception
\item Limited dataset size compared to RGB counterparts
\end{itemize}

\subsection{HSI Research and Applications}
\label{sec:Current State - HSI Applications}
Current research has identified several key areas where HSI can enhance or complement traditional RGB-based perception systems. While existing datasets mainly provide object-level annotations, there is significant potential to extend these applications toward material-based scene interpretation and detailed road surface analysis, both of which are critical for AD.

\subsubsection{Semantic Segmentation Evaluation}
\label{sec:Current State - HSI Semantic Segmentation}
Deep learning approaches for HSI analysis in ADAS/AD remain in their early stages, especially when compared to mature applications such as remote sensing. Table~\ref{tab:LitAndWorkOn_HSI} summarizes key methodologies applied to publicly available urban HSI datasets.

Early research primarily focused on dataset creation and fundamental evaluations, with limited application of deep learning methods. For example, initial studies like Basterretxea et al.\cite{basterretxea2021hsi} and Gutiérrez-Zaballa et al.\cite{gutierrez2022exploring} utilized per-pixel artificial neural networks (ANNs) and patch-based UNet architectures on HSI-Drive datasets.


Recent work (2024 onwards) has shifted toward multi-dataset benchmarking and more advanced architectures. Theisen et al.\cite{theisen2024hs3} proposed HS3-Bench, establishing standardized evaluation protocols and baseline performances, using DeepLabv3+\cite{chen2017deeplab} and UNet models across HyKo-VIS, HSI-Drive, and H-City datasets. Similarly, Shah et al.\cite{shah2024hyperspectral} evaluated multiple semantic segmentation networks on HSI data, including DeepLabv3+, HRNet\cite{sun2019high}, PSPNet~\cite{zhao2017pyramid}, and UNet, alongside attention-enhanced UNet variants using CBAM~\cite{woo2018cbam} and Coordinate Attention (CA)\cite{dang2021coordinate}. Other studies have explored multi-scale spectral feature extraction modules and dimensionality reduction techniques to better leverage hyperspectral information\cite{shah2025multiscalespectralattentionmodulebased,li2025csnrjmimbasedspectral}.

Despite these advances, several challenges remain: (1) The highly dynamic nature of driving environments complicates spectral feature extraction compared to relatively static domains like remote sensing~\cite{shah2024hyperspectral}; (2) Existing deep learning models are often not fully optimized to exploit the rich spectral signatures, performing better with conventional RGB data~\cite{shen4560035urban}; (3) A lack of large-scale, well-annotated HSI datasets limits the robustness and generalizability of current models~\cite{shah2024hyperspectral}.

\subsubsection{Material Discrimination and Surface Analysis}
\label{sec:Current State - Material Discrimination}
Material discrimination represents a complementary layer between scene-level semantic segmentation and object-level recognition in HSI-based perception. While semantic segmentation focuses on labeling regions and objects, material discrimination exploits HSI’s fine-grained spectral signatures to characterize surfaces and environmental conditions that are not easily distinguishable in RGB imagery.

Studies using the HSI-Drive dataset have demonstrated accurate identification of asphalt, water, glass, and various painted surfaces that appear visually similar in RGB~\cite{gutierrez2022exploring,PhotonfocusADAS}. This capability extends to road surface condition monitoring, where HSI can distinguish wet surfaces, detect ice formation, and classify pavement types~\cite{valme2024road}, which is critical information for assessing traction conditions and adapting vehicle behavior.

Such material-level understanding also benefits downstream perception tasks. For example, traversability estimation~\cite{jakubczyk2022hyperspectral}, enabling recognition of road surfaces such as ground, forest, asphalt, and grass. It also supports applications in automated road defect detection and infrastructure monitoring~\cite{abdellatif2019hyperspectral}, supporting both navigation safety and predictive maintenance.

\subsubsection{Object Classification and Recognition}
\label{sec:Current State - Object Detection}
HSI-based object classification research primarily targets challenging scenarios where RGB-based systems underperform. Early studies demonstrated HSI’s ability to separate pedestrians from complex backgrounds~\cite{herweg2012hyperspectral,herweg2013separability}, though these experiments were conducted in controlled settings rather than dynamic urban traffic. More recent work using the H-City dataset suggests that optimal spectral band selection can improve segmentation of pedestrians, cyclists, and urban infrastructure elements such as fences~\cite{li2025hyperspectralvsrgbpedestrian}, addressing important safety considerations for ADAS/AD systems.

HSI also shows promise under adverse weather conditions, where conventional RGB sensors suffer significant performance degradation. Preliminary results indicate that HSI maintains classification capabilities in low-light and rain-on-sensor scenarios~\cite{PhotonfocusADAS}, both of which can impair RGB-based perception. However, comprehensive comparative evaluations are needed to quantify these advantages and establish HSI’s role in robust, all-weather perception.

While current HSI-only approaches often underperform compared to RGB-based systems in semantic segmentation~\cite{shen4560035urban}. Recent advances, such as multi-scale spectral feature extraction modules, have started to show improved HSI performance~\cite{shah2025multiscalespectralattentionmodulebased}. Given the complementary nature of spectral and spatial information, HSI-RGB fusion strategies have strong potential to surpass the capabilities of individual modalities, enabling more reliable and adaptable perception systems for AD.

\subsection{Research Gaps and Challenges}
\label{sec:Current State - Research Gaps}
\noindent The transition from current limited HSI applications toward comprehensive perception systems for ADAS/AD faces several fundamental constraints that distinguish it from established remote sensing applications. While remote sensing benefits from mature processing pipelines and extensive datasets enabling techniques such as deblurring and denoising~\cite{li2010hyperspectral}, HSI development for automotive applications faces unique challenges.

\subsubsection{Dataset and Standardization Limitations}
Current HSI datasets for ADAS/AD suffer from several critical limitations: (1) Dataset scale remains insufficient, with the largest dataset (H-City) containing 1,330 annotated scenes compared to thousands in RGB datasets, like Cityscapes; (2) Annotation granularity varies significantly, with most datasets providing only object-level labels while material-based annotations remain sparse; (3) Spectral standardization is absent, preventing effective cross-dataset model generalization; (4) Class imbalance is severe, with road surfaces and vegetation dominating most datasets while critical objects, such as pedestrians represent less than 0.5\% of annotated pixels.

\subsubsection{Evaluation and Benchmarking Gaps}
The absence of standardized evaluation protocols represents a critical barrier to progress. Unlike RGB-based systems that benefit from standardized metrics and benchmarks, HSI research lacks consensus on performance evaluation. Studies employ different spectral band selections, varying class definitions, and inconsistent train-test splits, making cross-study comparisons unreliable and hindering systematic progress assessment.

\subsubsection{Algorithmic and Computational Challenges}
Technical challenges include: (1) HSI data requires significantly more computational resources than RGB equivalents: HSI-Drive's 25 channels represent 8× the data volume, while H-City's 128 channels represent 43× the data volume; (2) Limited architectural innovation, rather than exploiting spectral structure, most approaches adapt RGB-designed networks: U-Net, DeepLabv3+, etc; (3) Absence of effective multi-modal fusion strategies that combine HSI with established sensor modalities.

\subsubsection{Application-Specific Requirements}
ADAS/AD applications demand capabilities not addressed by current research: (1) Robust performance across diverse geographic regions and weather conditions; (2) Integration with existing automotive sensor architectures; (3) Compliance with automotive safety standards and certification requirements; (4) Cost-effective implementation compatible with mass-market vehicle constraints.

Addressing these limitations requires coordinated research efforts for dataset collection, algorithm development, hardware optimization, and standardization initiatives to realize HSI's potential for enhancing automotive perception. However, among these challenges, dataset standardization and scale represent the most immediate barriers to progress, while computational efficiency and multi-modal fusion require longer-term algorithmic optimization.

\begin{figure*}[htbp]
    \centering
    \includegraphics[width=0.99\textwidth]{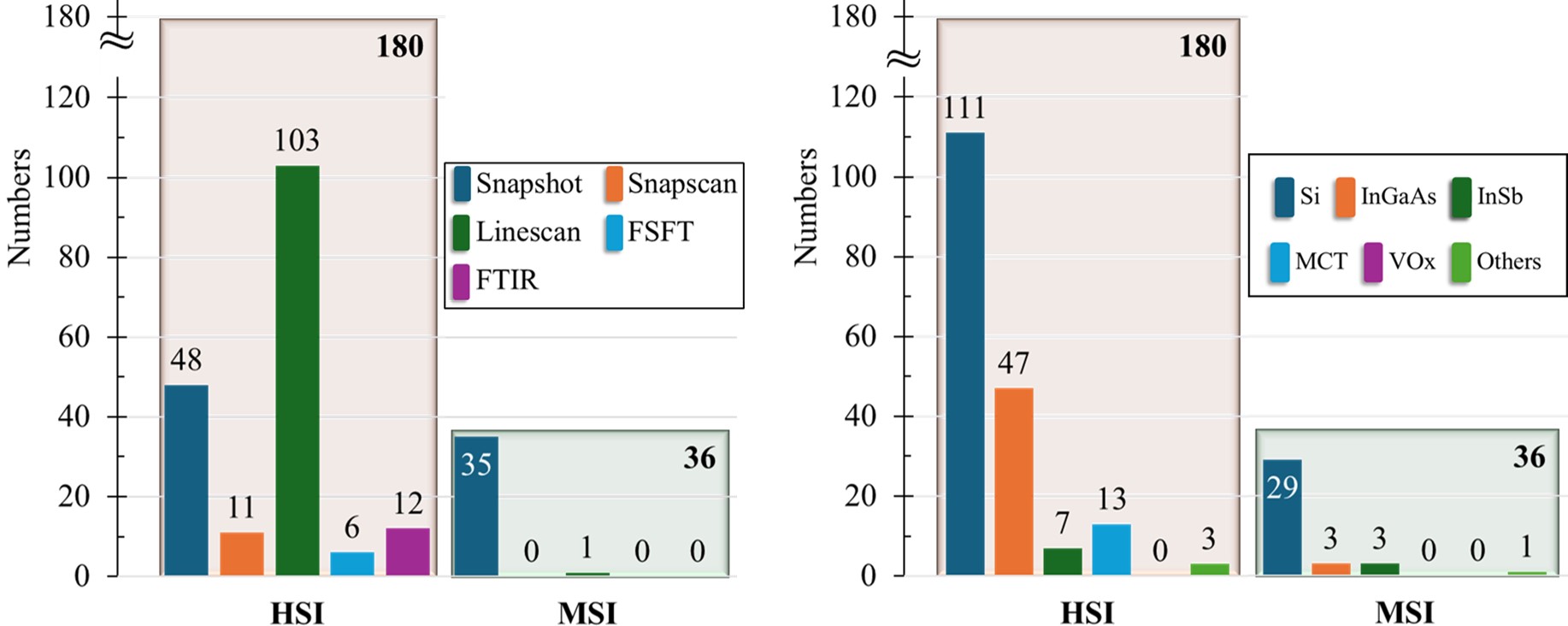}
    \caption{Distribution of imaging categories (i.e., HSI and MSI) based on (left) acquisition mode and (right) sensor type: Illustrating the number of cameras for each acquisition mode (Snapshot, Snapscan, Linescan, FSFT, FTIR) and sensor types (Si, InGaAs, InSb, MCT, VOx), respectively.}
\label{fig:AcquisitionAndSensorsByImagingCategory}
\end{figure*}

\section{HSI Sensors Availability and Suitability for ADAS/AD}
\label{Section:HSI_Availability_Suitability}
The commercial availability of hyperspectral imaging (HSI) sensors is a critical factor determining their practical deployment in automotive applications. The global HSI market demonstrates significant growth potential, with valuations reaching approximately USD 0.28-16 billion in 2023-24~\cite{BCCPublishing2025,SkyQuestt2025,MarketsandMarket2025,StraitsResearch2025,GrandViewResearch2025}, and projected compound annual growth rates of 8-12\% through 2030. This growth is driven by expanding applications in remote sensing, agriculture, and industrial inspection. In comparison, the automotive camera market achieved USD 3-9 billion in 2024~\cite{EmergenResearch2025}, indicating a substantial opportunity for HSI integration into vehicular systems. However, despite promising market trajectories, snapshot-based HSI cameras specifically designed for automotive deployment remain commercially scarce, creating a significant barrier to HSI adoption in ADAS/AD systems.

This section presents a comprehensive analysis of 216 commercially available HSI and MSI cameras, systematically evaluating their specifications against established ADAS/AD requirements. The assessment encompasses critical parameters, including spatial resolution, spectral coverage, frame rate, operating temperature, weight, and power consumption, all factors that directly impact the feasibility of automotive integration.

\begin{figure}[h!]
    \centering
    \includegraphics[width=0.485\textwidth]{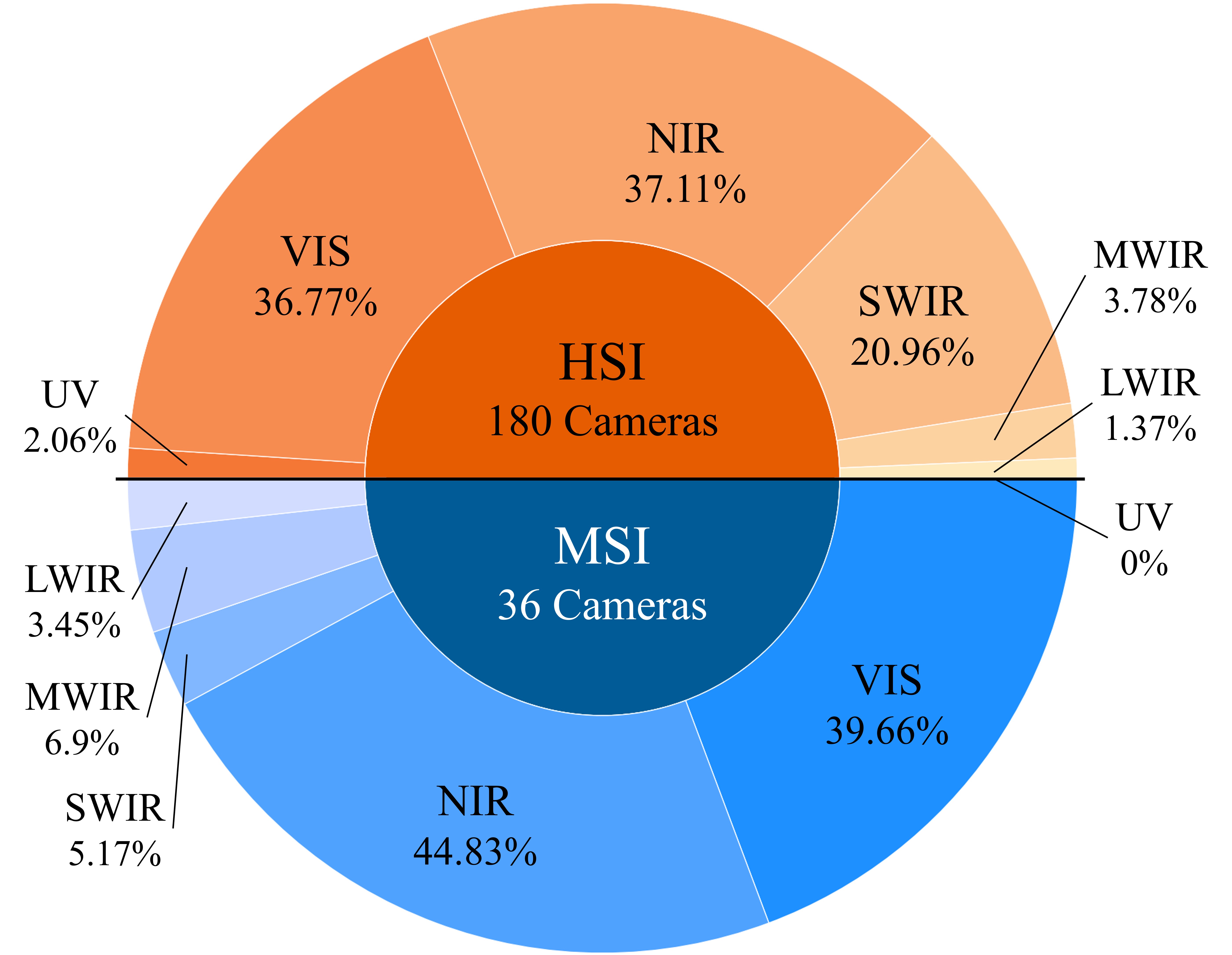}
    \caption{Distribution of spectral band coverage of commercially available HSI and MSI cameras reviewed in this study. The illustrated high alignment to the VNIR region demonstrates that current cameras are primarily designed for remote sensing applications and have limited availability for SWIR to LWIR applications required for adverse conditions and nighttime operations. Note: LWIR capabilities are limited to single-channel covering the broader 8-14$\mu$m thermal region. }
    \label{fig:HSIvsMSI_SpectralCoverage_Distribution}
\end{figure}

\subsection{Methodology and Evaluation Framework}
Our analysis employed a systematic market survey methodology, collecting specifications from major HSI/MSI manufacturers, including Specim, Headwall Photonics, IMEC, and others, between January 2024 and June 2025. The data, including spectral ranges, sensor types, and channel counts, was compiled directly from the manufacturers' and suppliers' official product datasheets and websites. It is important to note that these specifications are subject to change due to ongoing technological advances and product updates.

To establish automotive suitability, we defined a Matching Criteria Subset (MCS) based on fundamental ADAS/AD requirements:
\begin{enumerate}
    \item \textbf{Operating Temperature}: AEC-Q100~\cite{AEC-Q100_RevJ} compliance (minimum Grade 3: -40\textdegree C to +85\textdegree C)
    \item \textbf{Frame Rate $\ge$ 20 FPS}: Accommodating real-time applications with conservative refresh rate requirements, due to current HSI technology limitations
    \item \textbf{Spatial Resolution $\ge$ 1 megapixel (MP)}: Ensuring adequate scene detail with typical automotive cameras~\cite{sahin2019long}
    \item \textbf{Spectral Channels $\ge$ 30}: Meeting IEEE 4001~\cite{IEEE_P4001_2025} HSI definition
\end{enumerate}

These criteria reflect conservative estimates for general ADAS applications, recognizing that specific use cases may demand higher performance thresholds.

\begin{figure*}[h!]
    \centering
    \includegraphics[width=0.99\textwidth]{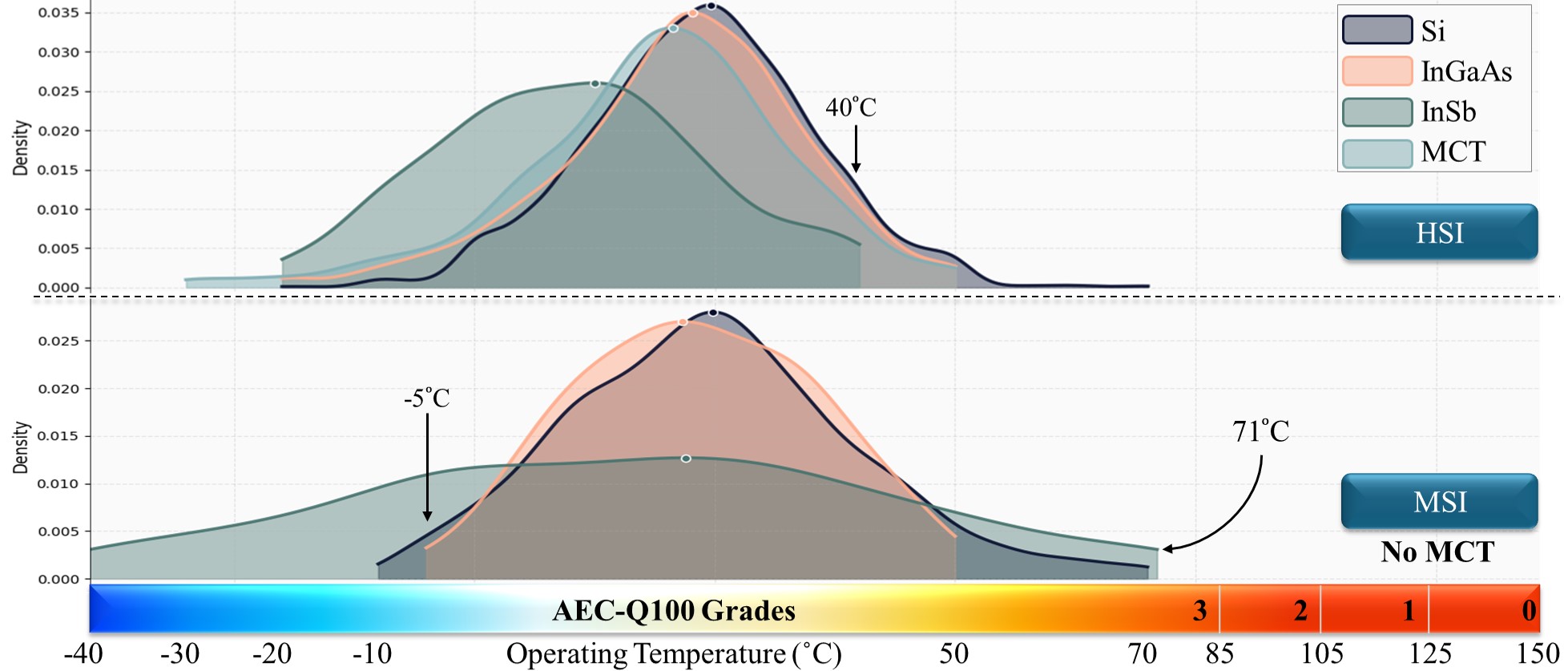}
    \caption{Operating temperature distribution of HSI and MSI cameras by sensor material in comparison to the Automotive Electronics Council's AEC-Q100 Standards. From -40\textdegree C to 85\textdegree C (Grade: 3), 105\textdegree C (Grade: 2), 125\textdegree C (Grade: 1), 150\textdegree C (Grade: 0). None of the reviewed cameras meet AEC-Q100 automotive qualification standards. Si and InSb sensors demonstrate wider operating temperature ranges compared to InGaAs and MCT.}
    \label{fig:HSIvsMSI_OperatingTemperature}
\end{figure*}

\subsection{Market Distribution and Sensor Technologies}
\subsubsection{Acquisition Methods and Sensor Materials}
The current HSI/MSI market is heavily skewed towards line-scan acquisition systems and the visible–near infrared (VNIR) spectral coverage, as shown in Fig.~\ref{fig:AcquisitionAndSensorsByImagingCategory}. Line-scan cameras constitute 47\% of available systems, reflecting their predominant use in industrial and remote sensing applications where continuous belt or platform motion enables spatial scanning. However, automotive applications require snapshot acquisition for instantaneous scene capture, significantly limiting suitable options.

Si sensors, as discussed in Section~\ref{Section:HSI_Fundamentals}(\ref{Section:HSI_Fundamentals-SUB:Sensor Technologies}), dominate the market, representing 65\% of available detectors, primarily operating in the VNIR region (0.4-1.1$\mu$m). InGaAs sensors extend coverage into the SWIR region (0.9-2.5$\mu$m), offering enhanced performance in adverse weather conditions, but requiring thermoelectric cooling and significantly higher costs. Together, Si and InGaAs sensors account for 87\% of commercial HSI cameras, as illustrated in Fig.~\ref{fig:AcquisitionAndSensorsByImagingCategory}.

The remaining market includes MCT and InSb sensors for MWIR and LWIR applications, though these typically require cryogenic cooling and present substantial integration challenges for automotive deployment.

\subsubsection{Spectral Coverage Analysis}

The utility of sensors is fundamentally determined by their spectral coverage, dictating the range of wavelengths they can detect. Different spectral bands provide unique information about objects and environments, crucial for tasks such as material identification, atmospheric correction, and enhanced visibility under various conditions.

Spectral band distribution, shown in Fig.~\ref{fig:HSIvsMSI_SpectralCoverage_Distribution}, illustrates the significant market concentration in the VNIR region. HSI cameras show 73\% of their spectral coverage concentrated in the VIS and NIR bands combined, while MSI cameras demonstrate an even stronger VNIR bias at 84\%. This distribution pattern reflects both the technological maturity of Si detector systems and the well-established applications in agriculture, remote sensing, and daylight imaging that drive market demand.

The limited availability of SWIR to LWIR capabilities presents critical gaps for the automotive domain. SWIR provides superior performance in fog, rain, and low-light conditions that are essential for all-weather ADAS applications. LWIR enables night-time pedestrian detection and object classification based on thermal signatures. Currently, LWIR-based cameras are limited to single-channel thermal cameras covering the broad 8-14$\mu$m window, lacking the spectral resolution for material discrimination.

These spectral coverage limitation constrains HSI applicability to primarily daytime and clear-weather scenarios, significantly reducing the technology's potential for comprehensive ADAS/AD deployment.

\begin{figure*}[h!]
    \centering
    \includegraphics[width=\textwidth]{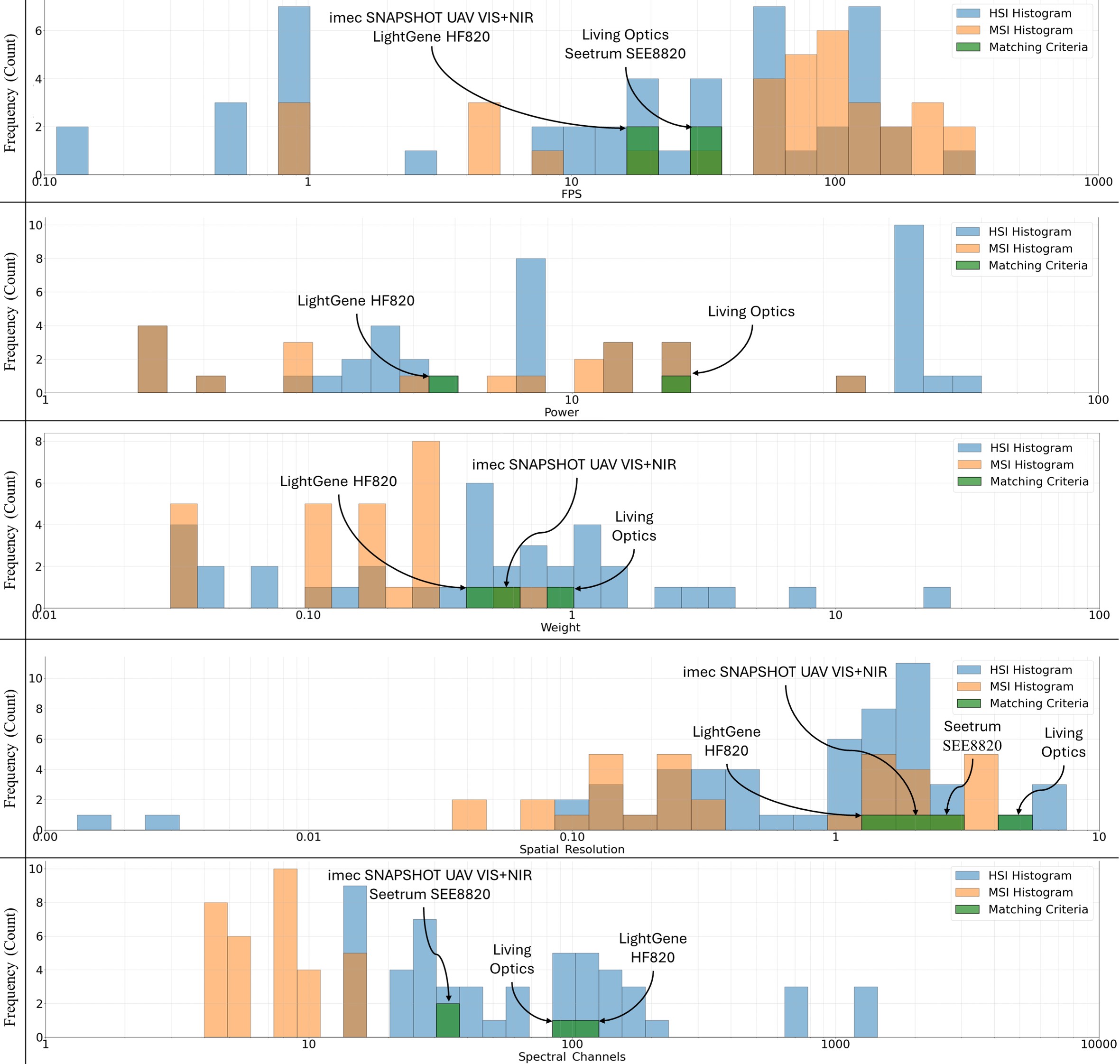}
    \caption{Performance characterization of commercially available and reviewed HSI (Blue) and MSI (Orange) cameras, crucial for ADAS/AD system applications. The histograms display the distribution of Frames Per Second (FPS), Power Consumption (in Watts: W), Weight (in kilogram: kg), Spatial Resolution (in megapixel: MP), and Spectral Channels among the reviewed cameras. Overlaid as green bars are the cameras that satisfy the criteria of FPS $\ge$ 20, spatial resolution $\ge$ 1 MP, and spectral channel $\ge$ 30. Out of 216 cameras, only four cameras (with their respective characteristics shown with arrows) meet these combined criteria, highlighting the limited availability and current landscape of high-performance, commercially viable solutions.}
    \label{fig:HSIvsMSI_WholeGrid}
\end{figure*}

\subsubsection{Operating Temperature}
Automotive environments demand exceptional thermal robustness, with components experiencing temperatures from sub-zero conditions to intense heat generated by engine compartments, solar loading, and geographic locations. The Automotive Electronics Council's AEC-Q100 standard defines qualification grades from Grade 0 to 3, as shown in Fig.~\ref{fig:HSIvsMSI_OperatingTemperature}, which illustrates the operating temperature distribution.

None of the 216 reviewed cameras meets even the basic AEC-Q100 Grade 3 automotive standards. Operating temperature analysis reveals significant variations among sensor technologies. Si-based cameras offer the broadest temperature tolerance, typically -20\textdegree C to +70\textdegree C for HSI, followed by InSb sensors with an operating range of -40\textdegree C to +71\textdegree C in MSI. Whereas, both InGaAs and MCT sensors exhibit more restrictive temperature ranges and typically require active thermal management systems to maintain optimal performance.

This overall non-compliance with automotive temperature standards represents a significant barrier to HSI deployment in ADAS/AD systems. While HSI and MSI cameras hold immense potential for enhancing ADAS/AD capabilities through advanced material identification, object classification, and environmental perception, their current operating temperature characteristics necessitate substantial sensor packaging and thermal management development.

\subsection{Snapshots' Performance Characteristics Assessment}
To assess the feasibility of HSI for ADAS/AD, a comparative analysis of key hardware and performance characteristics between HSI and MSI cameras is presented in Fig.~\ref{fig:HSIvsMSI_WholeGrid}. The figure summarizes the distributions of five critical parameters and overlays green bars to illustrate the cameras meeting the combined MCS, excluding operating temperature.

\subsubsection{Frame Rate (FPS)}
While general real-time ADAS applications typically require 25-30 FPS for dynamic scene analysis, current HSI technology limitations necessitated adopting a conservative 20 FPS threshold for the MCS criteria. Among the cameras surveyed, frame rates range from 0.11 FPS (9-second acquisition time) to 340 FPS (mean: 75.84 FPS), with HSI cameras averaging 61.89 FPS compared to 96.76 FPS for MSI systems. This performance gap reflects the computational overhead of increased spectral dimensionality. The MCS-compliant cameras operate between 20-30 FPS, confirming that high-performance HSI systems inherently trade frame rate for spectral resolution.

\subsubsection{Spatial Resolution}
Spatial resolution varies from 0.0013 to 7.5 MP, with a mean of 1.36 MP. While the 1 MP~\cite{sahin2019long} minimum threshold accommodates basic ADAS requirements, advanced applications such as long-range object detection and classification benefit from higher resolution. The four MCS-compliant cameras provide 1.56-2.5 MP resolution, with one system (LivingOptics) offering 5 MP but has only 4,384 spectral points uniformly spread over the spatial frame.

\subsubsection{Spectral Channels}
The number of spectral channels defines the richness of the spectral information captured. Across all data, spectral channels vary widely from 4 to 1440, with a mean of 114.28 channels. MCS-compliant cameras offer 31 to 125 spectral channels, with the LightGene HF820 providing the maximum channels.

\subsubsection{Power Consumption}
Power consumption spans 1.5-60 W across all reviewed cameras, with a mean of 15.41 W. HSI cameras on average consume significantly more power than MSI systems (18.56 W versus 8.66 W), driven by larger detector arrays and increased data processing requirements. This power consumption affects overall automotive integration by increasing thermal management complexity and electrical system loading requirements, especially when integrating multi-sensor suites. Power consumption for the MCS-compliant cameras ranges from 6-15 W, approaching acceptable automotive power budgets, but requiring careful thermal design.

\subsubsection{Weight}
Camera weight ranges from 0.03 to 27.2 kg (mean: 1.02 kg), with HSI averaging 1.58 kg compared to 0.20 kg for MSI cameras. Automotive integration requires the camera assemblies to be as light as possible, and the MCS-compliant cameras range from 0.5-1.0 kg, representing the upper limit of automotive acceptability.

\subsection{Challenges - Technological Gap Analysis}
Our comprehensive market analysis identified a limited number of HSI camera systems that currently satisfy the basic MCS criteria for automotive applications. The following camera systems demonstrate technical compatibility with our defined MCS requirements, listed in alphabetical order:
\begin{enumerate}
    \item \textbf{Imec Snapshot UAV VIS+NIR}: VNIR (0.46-0.9$\mu$m) is a dual camera-based HSI providing 31 spectral channels, 20 FPS, 2.2 MP resolution
    \item \textbf{LightGene HF820}: VNIR (0.45-0.95$\mu$m), 125 spectral channels, 20 FPS, 1.6 MP resolution
    \item \textbf{LivingOptics System}: VNIR (0.44-0.9$\mu$m), 96 spectral channels, 25 FPS, 5 MP resolution, but with uniformly spatially placed 4,384 spectral points
    \item \textbf{Seetrum SEE8820}: VNIR (0.38-0.98$\mu$m), 31 spectral channels, 30 FPS, 2.5 MP resolution
\end{enumerate}

\begin{figure*}[!h]
    \centering
    \includegraphics[width=0.99\textwidth]{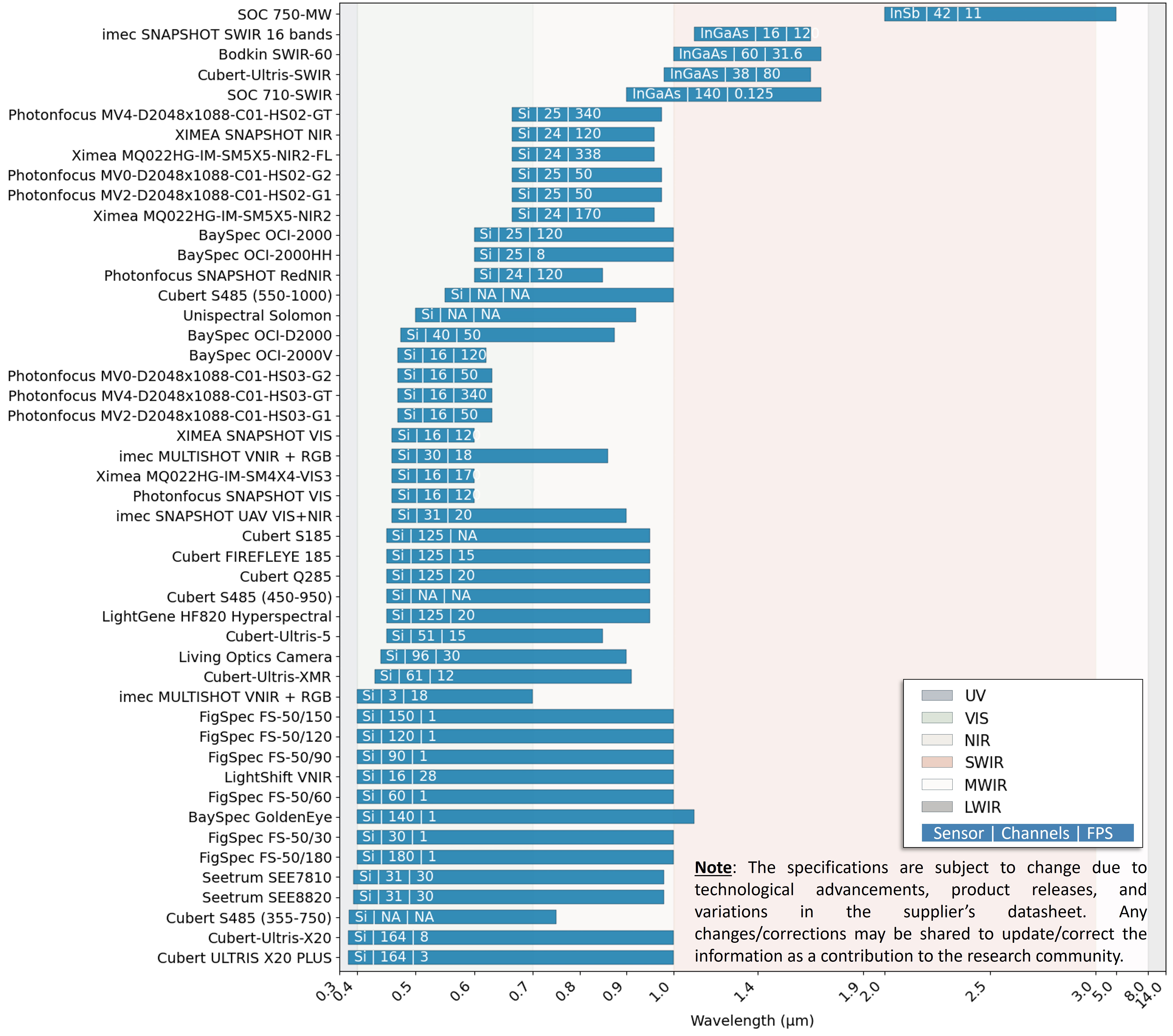}
    \caption{Overview of the reviewed commercially available snapshot HSI cameras only. The plot illustrates the wavelength range from 0.3 to 14$\mu$m (Not to Scale) with each camera covering its spectral range, categorized by sensor type (i.e., Si, InGaAs), number of spectral channels it captures, and FPS. The spectral coverage in corresponding spectral bands (e.g., VIS, NIR, etc) is also indicated. 'NA' denotes data not available in the camera datasheets or manufacturer's websites.}
    \label{fig:ShortlistedHSIcameraList}
\end{figure*}


While these systems represent significant technological achievements in snapshot HSI, they currently operate within the VNIR spectral region and require additional development to meet AEC-Q100 temperature qualification standards for automotive deployment. This highlights the existing technological gap between current commercial HSI capabilities and the comprehensive requirements for automotive integration. To assist researchers and practitioners in evaluating currently available snapshot HSI cameras according to their sensor specifications, spectral bands, spectral channels, and FPS performance, a comprehensive selection guide for snapshot HSI systems is provided in Fig.~\ref{fig:ShortlistedHSIcameraList}.


\section{Future Directions}
\label{Section:Future_Directions}
Building upon the challenges and limitations identified in our analysis of current HSI technologies and research, this section identifies some key future directions aimed at advancing the integration of HSI into ADAS/AD. We focus on addressing critical hardware constraints, expanding dataset diversity, enhancing deep learning models, and refining spectral rendering techniques to fully explore the potential of HSI for automotive applications.

\subsection{Development of Miniature Snapshot Cameras}
The integration of HSI technology into ADAS/AD systems requires overcoming critical hardware limitations identified in our analysis of 216 commercially available cameras:
\begin{itemize}
    \item Automotive-Grade Compliance: Achieving AEC-Q100 qualification is one of the primary barriers, as no reviewed cameras meet even Grade 3 standards (-40$\sim$+85\textdegree C). Si-based sensors show promise for temperature tolerance but require an extended spectral response beyond 1.1$\mu$m. SWIR capabilities require uncooled InGaAs arrays with improved temperature compensation.
    \item Miniaturization Requirements: Current HSI cameras require drastic hardware reduction through micro-optics, metalens arrays, and computational imaging. On-sensor processing using AI accelerators could enable real-time band selection and reduce data throughput.
    \item Extended Spectral Coverage: The predominant VNIR limitation (73\% of cameras) typically favors daytime and active sensor based applications, due to their limited performance in low light conditions. Future systems should investigate the potential of VNIR-to-MWIR capabilities through single or dual-band camera systems.
\end{itemize}

\subsection{Large and Diverse Datasets for Exploring HSI Potential}
\subsubsection{Dataset Diversity and Standardization}
A critical direction for future progress lies in the development of larger, more diverse, and standardized HSI datasets specifically designed for ADAS/AD research:
\begin{itemize}
    \item These datasets should aim to overcome the limitations of currently available datasets by incorporating a wider array of driving scenarios, including urban, rural, highway, and off-road environments.
    \item It is also essential to capture data under a broader spectrum of environmental conditions, encompassing various weather patterns (rain, snow, fog, sunny, cloudy), diverse lighting conditions (daytime, nighttime, dawn, dusk, glare), and seasonal variations across different geographical locations.
    \item These datasets should feature more comprehensive and fine-grained annotations, including detailed semantic segmentation for a larger variety of objects and materials relevant to autonomous driving, potentially extending beyond surface reflectance to include richer semantic labels.
    \item The adoption of standardized data formats, annotation guidelines, and evaluation metrics across the research community is strongly recommended to facilitate fair comparison of different algorithms and models and to ensure the reproducibility of research findings.
\end{itemize}

To evaluate HSI's potential in material classification and compare it with other RGB datasets, organizing data based on (1) basic material labels (glass, painted vs unpainted metal, concrete, and water), as in HSI-Drive, and (2) the H-City dataset's Cityscape-based annotations, offers a valuable framework for future dataset development.

\subsubsection{Potential Applications}
Future applications of HSI in ADAS/AD extend toward comprehensive scene understanding and predictive capabilities, subject to active research and comprehensive data availability, that include:
\begin{itemize}
    \item \textbf{Perception of Road and Weather Condition}: Advanced material characterization could enable vehicles to identify road traction by estimating surface friction properties through the detection of water and ice formation levels, and assess road surface stability before vehicle contact. The HSI's ability to detect material composition changes could enable early warning systems for road surface deterioration or infrastructure failures. In addition, better detection in adverse weather conditions~\cite{judd2019automotive}, such as fog, mist, haze, etc., where a sensor suite with SWIR sensitivity can surpass standard RGB-based perception systems. 
    \item \textbf{Sensor Fusion Applications}: HSI fusion with existing systems to create multi-modal perception frameworks and sensor calibration~\cite{st2017passive}, where spectral information could provide material context for LiDAR point clouds and validate radar reflections, potentially improving object classification accuracy in complex urban environments.
    \item \textbf{Night vision enhancement}: Detection in SWIR and MWIR regions by HSI sensor presents a promising night vision application, which could complement or even replace traditional thermal imaging systems. The enhanced spectral resolution could provide better object discrimination in low-light conditions while maintaining material identification capabilities.
    \item \textbf{Camouflaged and Background Blended Object Detection}: In addition to applications like enhanced pedestrian separability~\cite{herweg2013separability,herweg2012hyperspectral,li2025hyperspectralvsrgbpedestrian}, HSI's ability to detect distinct spectral signatures allows for the identification of objects camouflaged from RGB. This includes vehicles with unusual paint schemes, debris that blends with the road, and animals exhibiting natural camouflage. 
    \item \textbf{Real vs. Artificial or Fake Object Classification}: One potential application of HSI can be to counter the limitation of the metamerism effect to detect a real vs. fake object~\cite{song2024short}, for example, a real human vs. a billboard with a human posing image, or a real traffic light vs. a fake one printed on a t-shirt. This will highly favor reducing the false positives, eventually adding to the robustness of the road presence and environmental awareness.
\end{itemize}

\subsection{Deep Learning (DL) and Spectral-Spatial Models}
The effective processing and extraction of meaningful information from the high-dimensional data generated by HSI sensors for ADAS/AD tasks will crucially depend on the continued advancement and application of DL techniques, given the inherent complexity of HSI data analysis. Future research should prioritize the development of optimal spectral-spatial models capable of simultaneously leveraging both the material-wise spectral signatures and the spatial context of objects within the driving scene to achieve enhanced accuracy and robustness. This includes exploring the potential of hybrid models based on CNN and Transformers architectures for directly processing the hyperspectral data cube, as well as investigating the integration of attention mechanisms and graph neural networks to capture the intricate relationships between spectral and spatial features. Moreover, given the real-time processing requirements and resource limitations of onboard automotive computing systems for safety-critical applications, there is a significant need for research into the development of efficient and lightweight DL models that can be deployed on embedded automotive platforms with constrained computational resources. This may involve exploring techniques such as network pruning, quantization, and the utilization of specialized hardware accelerators like FPGAs.

\subsection{Color Science and Spectral Rendering}
The integration of HSI technology into ADAS/AD presents unique opportunities and challenges in color science and spectral rendering that extend far beyond traditional RGB imaging capabilities.
\begin{itemize}
    \item Illumination-Invariant Processing: Developing spectral descriptors that maintain material classification across varying automotive lighting conditions, such as solar angles, artificial lighting, mixed illumination, while preserving features critical for road surface analysis and object recognition.
    \item Adaptive Band Selection: Implementing context-aware algorithms that dynamically select optimal spectral bands based on environmental conditions and perception tasks, adapting processing for urban daytime versus nighttime highway scenarios.
    \item Cross-Modal Integration: Creating fusion techniques between HSI and existing RGB systems through spectral-to-RGB rendering for seamless pipeline integration and RGB enhancement using spectral information.
    \item Automotive Standards: Establishing spectral standardization protocols for consistent performance across regions, seasons, and atmospheric conditions, including automotive-specific color models that balance spectral fidelity with real-time computational requirements.
\end{itemize}

\section{Conclusion}
\label{Section:Conclusion}
This comprehensive review systematically evaluated HSI technology for ADAS/AD applications through analysis of over 216 commercially available cameras, existing datasets, and current research applications, establishing the first comprehensive baseline assessment of HSI capabilities for automotive deployment as of 2025. Our analysis reveals a fundamental disconnect between HSIs demonstrated potential and current commercial readiness for automotive applications. While HSI technology offers superior material discrimination and environmental characterization capabilities compared to conventional RGB cameras, enabling enhanced road surface classification, obstacle detection under challenging conditions, and material-based scene interpretation, critical barriers persist for widespread automotive adoption.

Only four of the 216 reviewed cameras meet basic automotive performance criteria ($\ge$ 20 FPS, $\ge$ 1 MP resolution, $\ge$ 30 spectral channels), and critically, none satisfy even the lowest AEC-Q100 automotive temperature standards (Grade 3: -40\textdegree C to +85\textdegree C). Market concentration in VNIR coverage (73\% of HSI cameras) further limits applicability to daytime or good lightning conditions scenarios.

Current HSI datasets suffer from insufficient scale (the largest contains 1,330 scenes versus thousands in RGB datasets), inconsistent spectral coverage, and limited environmental diversity. Algorithmic development remains limited, with most approaches adapting RGB architectures rather than exploiting HSI's spectral characteristics. Other limitations, like high computational overhead (8-43 times data volume versus RGB), power consumption (18.56 W average for available HSI cameras), and weight constraints (1.58kg average), present significant real-time deployment challenges for automotive systems.

Despite these barriers, the technology shows promise in material discrimination, adverse weather perception, and enhanced object classification. The rapid HSI market growth (8-15\% CAGR) suggests continued miniaturization of HSI technologies leading to automotive-grade solutions.

The future work should focus on (1) development of automotive-qualified snapshot cameras with extended spectral coverage; (2) large-scale, annotation-standardized, multispectral bands based HSI dataset creation; and (3) spectral-spatial algorithmic architectures optimized for real-time processing.

This review establishes the current state of HSI technology for automotive applications and provides essential guidance for advancing spectral imaging from laboratory demonstrations to production ADAS/AD systems. HSI success exploration requires coordinated efforts across hardware development, dataset standardization, algorithmic innovation, and automotive integration to bridge the identified gaps between HSI's potential and practical deployment requirements.

\bibliographystyle{IEEEtran}
\bibliography{ref}
\end{document}